# Artificial Intelligence-powered fossil shark tooth identification: Unleashing the potential of Convolutional Neural Networks

(Running title: SharkNet-X-CNN)

Andrea BARUCCI, Giulia CIACCI, Pietro LIÒ, Tiago AZEVEDO, Andrea DI CENCIO, Marco MERELLA, Giovanni BIANUCCI, Giulia BOSIO*, Simone CASATI and Alberto COLLARETA

A. Barucci, Istituto di Fisica Applicata "Nello Carrara", CNR-IFAC, Via Madonna del Piano, 10 - 50019 Sesto Fiorentino, Firenze, Italy; a.barucci@ifac.cnr.it

G. Ciacci, Istituto di Fisica Applicata "Nello Carrara", CNR-IFAC, Via Madonna del Piano, 10 - 50019 Sesto Fiorentino, Firenze, Italy; giuliaciacci8@gmail.com

P. Liò, Department of Computer Science and Technology, University of Cambridge, Cambridge, UK; pl219@cam.ac.uk

T. Azevedo, Department of Computer Science and Technology, University of Cambridge, Cambridge, UK; pl219@cam.ac.uk; tiago.azevedo@cst.cam.ac.uk

A. Di Cencio, Gruppo Avis Mineralogia e Paleontologia Scandicci, Piazza Vittorio Veneto 1, Badia a Settimo, 50018 Scandicci (FI), Italy & Istituto Comprensivo "Vasco Pratolini", via G. Marconi 11, 50018 Scandicci (FI), Italy & Studio Tecnico Geologia e Paleontologia, via Fratelli Rosselli 4, 50026 San Casciano Val di Pesa (FI); andrea.dicencio@gmail.com

M. Merella, Dipartimento di Scienze della Terra, Università di Pisa, via S. Maria 53, 56126 Pisa (PI), Italy; marco.merella@phd.unipi.it

G. Bianucci, Dipartimento di Scienze della Terra, Università di Pisa, via S. Maria 53, 56126 Pisa (PI), Italy & Museo di Storia Naturale, Università di Pisa, via Roma 79, 56011 Calci (PI), Italy; giovanni.bianucci@unipi.it

G. Bosio, Dipartimento di Scienze dell'Ambiente e della Terra (DISAT), Università di Milano-Bicocca, Piazza della Scienza 4, 20126 Milano (MI), Italy; giulia.bosio.giulia@gmail.com *corresponding author*

S. Casati, Gruppo Avis Mineralogia e Paleontologia Scandicci, Piazza Vittorio Veneto 1, Badia a Settimo, 50018 Scandicci (FI), Italy; sim.casati@gmail.com

A. Collareta, Dipartimento di Scienze della Terra, Università di Pisa, via S. Maria 53, 56126 Pisa (PI), Italy & Museo di Storia Naturale, Università di Pisa, via Roma 79, 56011 Calci (PI), Italy; alberto.collareta@unipi.it

KEYWORDS - AI; Elasmobranchii; Explainability; Palaeoichthyology; Shark Teeth; Taxonomic determination

ABSTRACT - All fields of knowledge are being impacted by Artificial Intelligence. In particular, the Deep Learning paradigm enables the development of data analysis tools that support subject matter experts in a variety of sectors, from physics and medicine up to the recognition of ancient languages. Palaeontology is now observing this trend as well. This study explores the capability of Convolutional Neural Networks (CNNs), a particular class of Deep Learning algorithms specifically crafted for computer vision tasks, to classify images of isolated fossil shark teeth gathered from online datasets as well as from the authors' experience on Peruvian Miocene and Italian Pliocene fossil assemblages. The shark taxa *that are included in the final, composite dataset (which consists of more than one thousand images) are representative of both extinct and extant genera, namely,* Carcharhinus, Carcharias, Carcharocles, Chlamydoselachus, Cosmopolitodus, Galeocerdo, Hemipristis, Notorynchus, Prionace *and* Squatina*. We developed, trained and tested a CNN, named SharkNet-X, specifically tailored on our recognition task, reaching a 5-fold cross validated mean accuracy of 85% to identify images containing a single shark tooth. Furthermore, we elaborated a visualization of the features extracted from images using the last dense layer of the CNN, achieved through the application of the clustering technique t-distributed Stochastic Neighbor Embedding (t-SNE). In addition, in order to understand and explain the behaviour of the CNN while giving a paleontological point of view on the results, we introduced the explainability method SHAP (SHapley Additive exPlanations), which is a game theoretic approach to explain the output of any Machine Learning model. To the best of our knowledge, this is the first instance in which this method is applied to the field of palaeontology. The main goal of this work is to showcase how Deep Learning techniques can aid in identifying isolated fossil shark teeth, paving the way for developing new information tools for automating the recognition and classification of fossils.*

INTRODUCTION

Applications of Artificial Intelligence (AI) are growing in popularity across a broad spectrum of scientific fields (Wang H. et al., 2023), from fundamental physics (Carleo et al., 2019; Ciacci et al., 2023), biology (Hassoun et al., 2021; Piazza et al., 2021; Richards et al., 2022) and healthcare (Barucci et al., 2021b; D'Andrea et al., 2023) to cultural heritage (Barucci et al., 2021a; Cucci et al., 2021; Bickler, 2021; Guidi et al., 2023). In the broad field of AI, Machine Learning (Domingos, 2012) and in particular Deep Learning (LeCun, 2015) algorithms are de facto driving the AI revolution by enabling researchers to harness the power of data, automate processes, improve decision-making and create innovative solutions that have a profound impact on society, economy and technology.

Palaeontology is starting to experience the same kind of cross-fertilization with AI, as highlighted by some recent works and projects (Liu & Song, 2020; Burton, 2022; MacFadden et al., 2023; Antonenko & Abramowitz, 2023).

For instance, in Perez et al. (2023), researchers developed an array of Machine Learning models capable of classifying images of extinct shark teeth, aiming to examine the benefits and drawbacks of Roboflow and Google's Teachable Machine, two free online platforms, as well as to identify techniques for enhancing paleontological datasets.

In Liu et al. (2023), three Convolutional Neural Networks (CNNs) were trained for taxonomic identification purposes with about 415 thousand photos from 50 different fossil clades, while in Ho et al. (2023) a CNN was introduced to hierarchically classify carbonate skeletal grains (both macro- and microscopically), thus enabling Linnaean taxonomic classification from a single image. A dataset of protoceratopsian dinosaur Computer Tomography pictures from the Gobi Desert (Mongolia) was developed by Yu et al. (2022) to evaluate the fossil segmentation capabilities of the Deep Learning autoencoder architecture U-net. A Deep Learning application for the identification of uncommon Cambrian microfossils was provided by Wang B. et al. (2022), whilst the identification of fossil brachiopods was the main emphasis of Wang H. et al. (2022). Work presented in Xu et al. (2020) employed Machine Learning to identify paleontological photomicrographs, while Liu & Song (2020) used CNNs to identify fossil and abiotic grains. Recently, Yu et al. (2024) examined over 70 paleontological AI investigations conducted since the 1980s, encompassing tasks like prediction, image segmentation, and the categorization of micro- and macrofossils.

Fostering the integration of AI techniques for data analysis in palaeontology is crucial due to the complexity of the field. The area of study at the intersection of these two disciplines may yield valuable insights from several perspectives. On the one hand, palaeontology can, for the first time, leverage AI-based powerful tools to examine data; on the other hand, AI can use paleontological data to build, test and refine its methods. Thus, palaeontology has all the potential to become one of the new frontiers in Artificial Intelligence applications today.

Here, we introduce a Deep Learning approach to the problem of the genus-level classification of images of isolated fossil shark teeth based on CNNs, a class of algorithms specifically tailored to deal with computer vision tasks. Thanks to the development of a specific dataset of shark tooth images, we were able to develop, train and test our own CNN, named SharkNet-X, tailored on the complexity of our problem, keeping into account the simplicity of the network architecture and ease of training. Our dataset allocates more than one thousand images, representing both extinct and extant shark genera. The dataset used to train SharkNet-X was built by merging publicly available fossilised shark teeth (sub)datasets along with images gathered by the authors based on their own experience with Peruvian Miocene and Italian Pliocene fossil materials.

The development and training of SharkNet-X allowed us to extract features from the tooth images contained in the dataset, which were then used to verify the presence of clusters among our data with the t-distributed Stochastic Neighbor Embedding (t-SNE) clustering technique (Van der Maaten et al., 2008; Kobak & Berens, 2019).

To the best of our knowledge, we introduced the idea of explainability of a CNN for the first time in the paleontological field by using the SHapley Additive exPlanations (SHAP) method (Lundberg et al., 2017; SHAP software; Azevedo, 2022). This powerful tool allows data scientists to gain insights into the factors driving predictions in Machine Learning models. Such an explainable approach to the results of the developed CNN allowed us to make a step towards bridging the gap between a pure Machine Learning approach and the paleontological approach of human taxonomists.

Our main goal here is to demonstrate the Deep Learning paradigm's ability to assist the identification of ancient shark teeth, setting the groundwork for the creation of new information tools for the automatic recognition and categorization of objects in the field of palaeontology.

FOSSIL SHARK TEETH AND THEIR USE IN SYSTEMATIC PALAEONTOLOGY

The marine vertebrates that comprise the class Chondrichthyes (also known as chondrichthyans, or cartilaginous fishes) belong in two subclasses: the poorly diverse Holocephali (currently consisting of a few more than 50 species of chimaeras) and the much more speciose Elasmobranchii, which in turn include the extant superorders Selachimorpha (more than 500 extant species of sharks in eight orders) and Batoidea (almost 700 extant species of rays in four orders) (Serena et al., 2020, and references therein) (Fig. 1). The extant chondrichthyan diversity is still only partially known, with new species being described on a yearly basis (White et al., 2023). The cartilaginous fishes have roamed the Earth's global ocean for more than 400 million years (Andreev et al., 2022), leaving a vast fossil record in their wake. Most of this record is represented by dental and dermal remains: this is due to the fact that teeth and scales are heavily mineralized in chondrichthyans, unlike the endoskeleton, which is largely cartilaginous. In particular, the elasmobranch teeth are coated with a stiff, strong, tough layer of enameloid (a calcium phosphate material; Wilmers et al., 2021) that makes them good candidates for being preserved as fossils. Further enhancing the fossil record of the elasmobranchs is their peculiar pattern of dental replacement, whereby their teeth are shed rapidly and replaced continuously throughout life (Kemp, 1999; Gillis & Donoghue, 2007; Tucker & Fraser, 2014; Berkovitz & Shellis, 2016), so much so that some extant sharks can loss thousands or even tens of thousands of teeth during their lifetime (Brisswalter, 2009). Luckily for palaeontologists, the shark dentitions often exhibit diagnostic morphological characters, hence the utility of isolated fossil shark teeth for taxonomic identifications to the genus or even species level (e.g., Sáez & Pequeño, 2010; Cappetta, 2012; Pollerspöck & Straube, 2018). Ultimately, the painstaking palaeontological study of the fossil shark tooth records allows for reconstructing a comprehensive and detailed picture of the evolutionary history and palaeobiodiversity of one of the most charismatic groups of marine vertebrates (Cappetta, 2012).

## DATASET OF SHARK TOOTH IMAGES

The dataset developed in this work to train and test the CNN SharkNet-X was built on the basis of eight separate (sub)datasets. Six such (sub)datasets are publicly accessible and contain labelled pictures of shark teeth coming mostly from palaeontological collections of the Calvert Marine Museum and Florida Museum of Natural History, while the remaining two were created by the authors based on the palaeoichthyological collection of the G.A.M.P.S. (Gruppo A.V.I.S. Mineralogia Paleontologia Scandicci) permanent exhibition (Badia a Settimo, Scadicci; Italy) (e.g., Cigala Fulgosi et al., 2009; Collareta et al., 2018, 2020) as well as on the authors' first-hand experience on Peruvian Miocene fossil assemblages from the East Pisco Basin stored at the Museo de Historia Natural de la Universidad Nacional Mayor de San Marcos (Lima; Peru) (e.g., Landini et al., 2017, 2019; Collareta et al., 2021; Bosio et al., 2022). Tab. 1 reports information about each (sub)dataset, while some examples of the tooth images included therein are shown in Fig. 2.

In order to build a dataset suitable to train a CNN, some steps are necessary. Firstly, all the images comprising the eight (sub)datasets were merged into a single composite dataset containing more than 1600 images. Then, a deep manual check was performed, during which duplicate images and photos of taxa represented by very few specimens were removed. Additionally, photos displaying hand-held or weirdly oriented teeth as well as severely damaged specimens were also eliminated. The resulting dataset numbers about 1300 pictures of teeth in labial or lingual view. All these images underwent a pre-processing step, including the removal of the background to exclude associated elements such as museum labels and scale bars, and the creation of a black uniform background. Examples of the resulting pictures are shown in Fig. 2. Teeth are depicted in either lingual or labial view, mostly with their crown pointing downward, but some images with upward-directed main cusps are also present.

The shark taxa allocated in the final dataset include both extinct and extant genera, namely: *Carcharhinus*, *Carcharias*, *Carcharocles*, *Chlamydoselachus*, *Cosmopolitodus*, *Galeocerdo*, *Hemipristis*, *Notorynchus*, *Prionace* and *Squatina.* Tab. 2 reports on the number of images available for each genus. This image collection helps to increase the generalizability of our model by taking into consideration variations in terms of size, orientation, light condition, and intrageneric variability (with some relevant exceptions such as the monotypic *Prionace*, most of the studied genera are represented by more than a single species in our dataset).

It is also important to keep in mind that some of the considered genera, including *Hemipristis* and *Notorynchus*, are characterized by a high degree of heterodonty (either monognathic or dignathic), which means that major morphological differences exist between the upper/lower and/or anterior/posterolateral teeth, thus further complicating their automated identification.

# SHARKNET-X: A CONVOLUTIONAL NEURAL NETWORK FOR SHARK TEETH CLASSIFICATION

*SharkNet-X architecture*

SharkNet-X is a Convolutional Neural Network that was implemented in Python, using Keras (Chollet, 2021; Keras, 2015) and Tensorflow. It consists of convolutional (2D), max pooling, flatten, drop-out and dense layers (Abadi et al., 2015). Tab. 3 reports the network's architecture, layer by layer, showing the layer type, the output shape and the number of parameters. A sufficiently small size of 224 × 224 pixels has been empirically selected for the input images, which demonstrated to lessen the computational load without compromising the classification results while enabling easy detection of the shark tooth characteristics.

SharkNet-X architecture described in Tab. 3 was empirically derived by exploring different hyperparameter configurations. Specifically, we explored random selection of batch sizes (i.e., 16, 32, 64, 128), number of epochs (from 10 to 100) and initial learning rates (0.01, 0.001, 0.0005, 0.0001). We divided the dataset into training, validation, and test sets (see below for details), where the best hyperparameters were selected given the best performance on the validation set. Best performances were achieved with the Neural Network trained using the Adam optimiser (Kingma, 2014) with an initial learning rate of 0.0001, batch size of 128, 60 epochs, and sparse categorical cross-entropy loss. Early stopping was used during training to prevent overfitting. The CNN model, which has been fine-tuned in terms of architecture and hyperparameters, will henceforth be denoted as best model. Accuracy, balanced accuracy and other metrics were used to measure the performance of the CNN.

*Dataset for training*

In order to properly train the network and assess the model generalisation capability, the dataset was used in two different ways. The first approach involved using all the available images to assess SharkNet-X mean performance (and corresponding standard deviation). This was achieved by employing a 5-fold Cross-Validation scheme, as outlined in studies by Ng (1997), Hastie (2009), and Raschka (2019). The second approach adopted a different strategy by splitting the data into three subsets: training, validation and test sets. This enabled the training and evaluation of the final model on distinct sets of data.

It is important to outline this concept of training, validation and test sets, because it is at the root of all Machine Learning models developing. During the training of a Neural Network, the dataset is typically divided into two main subsets: the training set and the validation set. Within the context of Supervised Learning (Bishop 2006), the training set is a subset consisting of input data along with corresponding target labels (or values in the case of regression tasks). The model learns from the training set by adjusting its parameters (e.g., weights and biases) through optimization algorithms like gradient descent to minimise the loss function. The training process involves iterating through the training set multiple times (epochs), updating the model parameters to improve its performance in making predictions.

The validation set is a separate subset of the dataset used to evaluate the performance of the model during training. It serves as an independent sample to assess how well the model generalises to unseen data. The validation set helps monitor the model performance, detect overfitting (where the model learns to capture noise or irrelevant patterns from the training data rather than generalise, resulting in poor performance on unseen data), and tune hyperparameters. The model performance on the validation set is evaluated periodically during training, typically after each epoch, to determine if further training is improving or worsening the model ability to generalise (Bishop & Bishop, 2023). The train/test division was done with a ratio of about 90/10 (see Tab. 2).

It is important to stress that this concept of separating the dataset into these two subsets helps ensure that the model learns effectively and generalises well to new, unseen data. Ensuring this ability is a key aspect when developing a Machine Learning model. Returning to the first approach, Cross-Validation is another well-established technique used in Machine Learning to evaluate the performance of a model by partitioning the dataset into subsets, training the model on some of the subsets, and testing it on the remaining subset.

This process is repeated multiple times, with different partitions used for training and testing each time. Cross-validation helps assess the generalisation ability of the model and reduces the risk of overfitting.

*Feature clustering*

The employment of clustering methods applied to features extracted by CNNs has garnered increasing attention due to their potential in revealing underlying structures within data (Guérin et al., 2021). Features extracted by a CNN through its convolutional layers capture hierarchical representations of input images, while the last dense layer provides a high-dimensional feature vector that can be used for clustering similar images together, thus enabling the unsupervised analysis and organization of image datasets.

Here, we employed an unsupervised clustering method called t-SNE to the last dense layer of SharkNet-X. t-SNE is a popular dimensionality reduction technique used in Machine Learning and data visualization. It aims to map high-dimensional data points into a lower-dimensional space, typically 2D or 3D, while preserving the local structure of the data as much as possible. t-SNE works by modelling the similarity between data points in the original high-dimensional space and then representing these similarities in the lower-dimensional embedding. This helps to reveal clusters and patterns in the data that might not be easily discernible in the original high-dimensional space.

Exaggeration and perplexity are two crucial t-SNE hyperparameters that affect the final visualization, influencing how the algorithm balances between capturing local and global structures in the data. While perplexity controls the effective number of neighbours considered for each data point, exaggeration enhances the separation between clusters in the lower-dimensional embedding space, making them more visually distinct. A low perplexity value causes t-SNE to focus more on local structure, while a high perplexity value considers a larger neighbourhood for each data point. Typically, perplexity values in the range of 5 to 50 are used, while it is a common practice to use exaggeration with a factor of 4 or 12. It is often necessary to experiment with different values of perplexity and exaggeration to empirically find the optimal combination that reveals the structure of the data.

## CLASSIFICATION RESULTS

SharkNet-X was trained from scratch, using all images (1309) with the two approaches introduced above, Cross-Validation and best model, while performances were measured using accuracy, balanced accuracy and other metrics.

Accuracy in particular measures the proportion of correctly predicted instances out of the total number of instances in a dataset, while balanced accuracy adjusts the score to account for imbalanced datasets, where the number of instances in different classes is not evenly distributed. In order to do so, it calculates accuracy for each class individually and then averages these accuracies across all classes, providing a more reliable measure of performance on imbalanced datasets. In Neural Network training, validation, and testing, balanced accuracy helps ensure that the model performs well across all classes, not just the most populated class.

Another important parameter to look at when training a Neural Network is the Loss function, which quantifies how well the model's predictions match the true values of the target variable. Loss is typically calculated using a function, such as categorical cross-entropy for classification tasks or mean squared error for regression tasks. The goal during training is to minimise the loss function, which involves adjusting the model parameters (e.g., weights and biases) through optimization algorithms like gradient descent. Lower loss values indicate better agreement between the model's predictions and the true values, reflecting improved performance.

These metrics are essential for assessing the performance and effectiveness of Neural Networks during the training, validation and testing phases, providing valuable insights into the model's behaviour and capabilities.

*Cross-validation*

SharkNet-X obtained an overall validation accuracy of 83.5 ± 1.0% and a test accuracy of 84.8 ±2.6 %. Among the 5 folds, the model reached best (highest) validation and test accuracy scores of 84.81% and of 87.96%, respectively. Results for each of the 5 folds are reported in Tab. 4, while relative learning curves are shown in Fig. 3.

*Best model*

SharkNet-X was subsequently tested on a single train/test split of the images, with a splitting ratio of 90/10 for a total of 1299 images. In this case, we obtained a prediction accuracy of 88% and a prediction balanced accuracy of 83%. All the prediction metrics used to assess the performance of SharkNet-X are reported in Tab. 5. Learning curves for accuracy and loss are shown in Fig. 4, while the confusion matrix for the test set is shown in Fig. 5.

*Unsupervised dense layer feature clustering*

We used the last dense layer of the CNN to extract 128 features for each image of the dataset. After standardization, the resulting feature data matrix (1309 images × 128 features) was given as input to the t-SNE algorithm implemented with SciKit-Learn (Pedregosa et al., 2011). Results are shown in Fig. 6 for the combination of hyperparameters given by a perplexity of 30 and an exaggeration of 12.

## CNN EXPLAINABILITY

Explainability of the results of a Neural Network is one of the most important and complex aspects of Deep Learning. However, this step is necessary in order to move these algorithms into applications in science, giving to the experts in the field not just a classification score, but an explanation of why we are obtaining these results. Explainability can also provide a feedback mechanism, where experts can help to improve the Deep Learning model looking at the explained results and, if necessary, contribute to modify the algorithm architecture or training. In our case, we implemented SHAP - SHapley Additive exPlanations, an algorithm based on Shapley values, a concept from cooperative game theory. Shapley values have been introduced in 1953 by Lloyd Shapley (Shapley, 1953), and have been used in the past to compute explanations on model predictions (Lipovetsky & Cocklin, 2001; Štrumbelj, 2014; Owen, 2017).

SHAP is able to explain the output of any Machine Learning model, quantifying the contribution of each feature to the prediction made by the model, thus providing insights on how the model arrived at a particular prediction. The core idea behind SHAP is rooted in cooperative game theory, specifically the concept of Shapley values, which assigns a value to each player in a cooperative game based on their contribution to the overall outcome. SHAP computes Shapley values for each feature, indicating how much each feature contributes to the difference between the actual prediction and the average prediction. SHAP values then provide interpretable explanations for individual predictions, highlighting which features pushed the prediction towards a certain outcome and which features pulled it away. These explanations can help users understand and trust complex Machine Learning models, identify influential features, detect biases, and debug model behaviour.

In the context of palaeontology, SHAP can help fill the gap between a pure data driven approach to classification and the necessity of a paleontological explanation of the results obtained by the model. In the case of images, which is relevant to our task of automating the classification of shark tooth photographs, SHAP provides insights on why a particular image was classified in a certain way by attributing the contribution of each pixel to the model decision. SHAP analyses the importance of individual pixels or regions of the image in influencing the model prediction. It computes SHAP values for each pixel, indicating how much a very pixel contributes to the difference between the model's prediction for the image and its average prediction. Positive SHAP values indicate that a pixel contributes to a higher prediction score for a particular class, while negative values indicate the opposite. By visualising these SHAP values, users can gain insights into which parts of the image are most influential in the model decision-making process. This can help in understanding why the model classified the image in a certain way, identifying important features or patterns in the image, and assessing the model strengths and weaknesses. Overall, SHAP provides a powerful tool for interpreting and explaining the decisions made by image classification models, enhancing transparency, trust, and understanding of these models in various applications.

We applied SHAP to the classification results of SharkNet-X, obtaining a heatmap of the SHAP values for image inputs of representative examples of each genus (Fig. 7). This heatmap provides a visual representation of the importance or contribution of each pixel to the model prediction for a particular image. A colour gradient is used to indicate the magnitude of the contribution, with warmer colours (e.g., red) representing higher or positive contributions and cooler colours (e.g., blue) representing lower or negative contributions (see Figs 7, 8). The heatmap often reveals localised patterns or regions of interest within the image that are particularly influential in the model decision-making process. These regions may correspond to specific features, objects or structures that are relevant to the classification task. Overall, the heatmap generated by SHAP provides valuable visual insights into the inner workings of image classification models, helping users to interpret and understand model predictions in a transparent and interpretable manner.

The warranted consistency of SHAP values help explain why it gained so much popularity in recent years. Indeed, these and other advantages have been highlighted in recent literature reviews (Arrieta, 2020; Bodria, 2023; Heuillet, 2021; Vilone, 2021). SHAP is presented by Linardatos et al. (2020) as "the most complete method, providing explanations for any model and any type of data, doing so at both a global and local scope", and "[together with LIME (Ribeiro 2016)], by far, the most comprehensive and dominant across the literature methods for visualising feature interactions and feature importance".

An additional reason for choosing CNNs rather than Transformers (Vaswani, 2017), an emerging Deep Learning architecture that is increasingly popular and among the highest performing, is related to the fact that while significant progress has been made in the field of explainable artificial intelligence (XAI; Samek, 2017), a notable gap remains in effective solutions for explaining transformer-based models. Transformers, despite their superior performance in various natural language processing tasks, pose unique challenges for explainability due to their complex architecture and attention mechanisms. Current approaches for XAI in transformer models often lack interpretability or fail to provide comprehensive insights into model decisions.

As a result, much of the research in XAI continues to rely on CNNs for their comparatively more interpretable representations and established methods for explainability (Zeiler, 2014). CNNs have been extensively studied and have well-established techniques such as feature visualization, occlusion analysis, and saliency mapping, which facilitate understanding of model behaviour and decision-making processes. While efforts are underway to develop effective XAI techniques for transformers, the field still predominantly relies on CNN-based approaches for their practical utility and interpretability.

DISCUSSION

The cross-validation and best model results confirm the good generalizability and prediction performances of SharkNet-X, respectively.

Results for the test set using the best model are summarised by the Confusion Matrix shown in Fig. 5, which shows that the most frequent mispredictions regarded 1) teeth of *Cosmopolitodus* being misinterpreted as belonging to *Carcharhinus*, and 2) teeth of *Carcharhinus* being misinterpreted as belonging to *Hemipristis*. The former issue may be due to the fact that some upper teeth of broad-toothed *Carcharhinus* species (i.e., the 'bull group' sensu Cappetta, 1987) are superficially reminiscent of the upper teeth of *Cosmopolitodus* in terms of overall outline, absence of cusplets and relatively short root lobes, although major differences do obviously exist (e.g., in terms of vascularization pattern and presence/absence of serrations). The misinterpretation of some teeth of *Carcharhinus* as belonging to *Hemipristis* may also be due to the first-order similarities that exist between the upper, cutting-type teeth of *Hemipristis* and the corresponding teeth of broad-toothed *Carcharhinus* species.

The high-dimensional representation learned by the last dense layer of the CNN is confirmed by the 2D t-SNE visualisation shown in Fig. 6. The extracted features exhibit clustering patterns corresponding to the shark genera, indicating the effectiveness of the CNN in capturing meaningful distinctions within the dataset. The best individualized cluster appears to be the one concerning *Chlamydoselachus*. This is not surprising as the tricuspid teeth of *Chlamydoselachus* are by far the most distinctive (in other words – 'unmistakable') of the whole dataset. Conversely, the cluster that describes the genus *Carcharhinus* partly overlaps with a few others. Once again, this comes as no surprise considering that *Carcharhinus* spp. are characterized by a cutting-clutching dentition wherein the upper and lower teeth are often different from each other in general morphology (e.g., Bourdon, 1991), and that different species of *Carcharhinus* (including both narrow- and broad-toothed forms) are present in our dataset, which concur to create a wide intrageneric variability in terms of tooth shape. Numerousness may also play a role here, as *Chlamydoselachus* and *Carcharhinus* are the least and most populated genera in our dataset, respectively.

SHAP results are shown in Fig. 7 for some representative examples of each genus. In the figure, SHAP values are given for each example and for each possible predicted outcome. Generally speaking, the distribution of warm- and cool-coloured pixels indicate that the characters that the CNN keeps as most diagnostic cluster along and inside the tooth margins, suggesting that it is the tooth outline (and especially the crown outline) that leads the process of automated identification. Thus, it is not surprising that clusters of genera characterized by similar dental crown morphologies are highlighted by the SHAP results. Considering for a moment the case of *Prionace*, similarities are indicated by the SHAP results between the upper teeth of *Prionace* and *Carcharhinus* (Figs 7, 8). Here, the CNN is in good company, as *Prionace* is now known to have evolved from within the genus *Carcharhinus* (Compagno, 1988; Dosay-Akbulut, 2008; Dunn et al., 2020).

Although some features of the SHAP results are not readily interpretable, they are often instructive. For example, features of *Hemipristis* are found in the three main cusps that feature prominently in images of *Chlamydoselachus* (Fig. 8). It is tempting to hypothesize that the CNN recognizes characters of the clutching-type, often cusplet-bearing crowns of lower teeth of *Hemipristis* in each of the three main cusps of *Chlamydoselachus*, which in turn evokes Herman et al.'s (1993) hypothesis on the evolutionary origin of the teeth of the latter genus: "Three separated (principal) cusps, with a small cusplet in between them, as well as the broad interspaces between the tooth rows indicate a possible ancient development of merging originally three separate tooth rows into one in *Chlamydoselachus anguineus*." Furthermore, that the CNN recognized features of *Galeocerdo* in teeth of *Notorynchus* (Fig. 8) is not surprising, as the former genus is characterized by teeth with a distally deflected main cusp followed distally by a shoulder comprised of progressively smaller cusplets that recall in some way the comb-like design of the latter's teeth.

From a data analysis point of view, it is important to outline some characteristics of the obtained SHAP heatmaps. As mentioned above, warm- and cool-coloured pixels (corresponding to higher and lower SHAP values, respectively) are generally found along or inside the tooth borders. In some cases, however, we found such pixels outside the main region of the image. This outcome is to be expected, as this method focus exclusively on explaining what the CNN is "looking at" in the input image in order to discriminate different genera of shark teeth.

LIMITATIONS OF THIS STUDY AND FUTURE PERSPECTIVES

Although the performances achieved by SharkNet-X are good, the clustering of t-SNE meaningful and the results of SHAP quite impressive, it is important to point out the limitations of our work in order to indicate future improvements and advancements. Firstly, one of the limitations consists in the small number of specimens included in some of the considered genera (e.g., *Chamydoselachus, Cosmopolitodus* and *Squatina*). Addressing this challenge, e.g. via the implementation of data augmentation techniques (Shorten, 2019; Mumuni, 2022) might result in improving the model's performances. At the same time, the SharkNet-X architecture itself could be improved, or we could leverage on transfer learning approaches (Zhuang, 2020).

It is also worth noting that a more systematic analysis will be necessary in future works to assess the consistency of the t-SNE results due to the stochasticity and hyperparameter dependency of the method.

Another limitation concerns the interpretability of the results provided by SHAP. While the SHAP heatmaps confirm that the CNN is "looking at" the right places (i.e., the tooth borders and inner regions), the method does not explain how this information is used by the network in order to perform the prediction. This limitation is inherent to the SHAP post-hoc approach to Explainability.

By acknowledging these limitations, we pave the way for future research endeavours aimed at addressing these challenges.

CONCLUSIONS

We explored the efficacy of a Convolutional Neural Network in the genus-level classification of fossil shark teeth. The proposed network, SharkNet-X, demonstrates promising performances, achieving a prediction accuracy up to 88%. Additionally, the t-SNE clustering analysis conducted on the last dense layer of the CNN reaffirmed the quality of the learned representation, clearly demonstrating distinct clustering patterns corresponding to different genera. Hopefully, this model will serve as an initial step toward the creation of research software dedicated to specimen classification and the analysis of tooth morphology.

In addition, we introduced for the first time in the field the application of an explainability method for CNN. This method, SHAP, allowed us to obtain a visual heatmap to interpret the output of the CNN. Such maps were investigated by palaeontologists in order to understand what kind of patterns/features the neural network was looking at during the identification process.

Although certain limitations persist, we are confident that future research will significantly advance our comprehension and utilization of Deep Learning techniques in the field of shark tooth classification and beyond.

ACKNOWLEDGMENTS

Authors gratefully acknowledge J. Agresti, C. Cucci, G. Magni, D. Mugnai, F. Rossi, S. Siano and P. Di Maggio from CNR-IFAC, S. Fiore (CNR), C. Canfailla and F. Argenti from DINFO-UniFI. A special thanks to A. M. Addabbo, S. Taruffi and to all the students from the "Istituto di Istruzione Superiore Tecnica e Liceale Russell Newton" (Scandicci, Italy), who supported the G.A.M.P.S. dataset creation. Last but not least, we thank the many researchers who made publicly available the six tooth image datasets used in this work.

M.M., G.Bi., G.Bo. and A.C. acknowledge financial support under the National Recovery and Resilience Plan (NRRP), Mission 4, Component 2, Investment 1.1, Call for tender No. 104 published on 02/02/2022 by the Italian Ministry of University and Research (MUR), funded by the European Union – NextGenerationEU – Project Title: BIOVERTICES (BIOdiversity of VERTebrates In the CEnozoic Sea) – CUP I53D23002070 006 – Grant Assignment Decree No. 965 adopted on 30/06/2023 by the Italian Ministry of Ministry of University and Research (MUR).# REFERENCES

Abadi M., Agarwal A., Barham P., Brevdo E., Chen Z., Citro C., Corrado G.S., Davis A., Dean J., Devin M., Ghemawat, S., Goodfellow, I., Harp, A., Irving, G., Isard, M., Jia, Y., Jozefowicz, R., Kaiser, L., Kudlur, M., Levenberg, J., Mane, D., Monga, R., Moore, S., Murray, D., Olah, C., Schuster, M., Shlens, J., Steiner, B., Sutskever, I., Talwar, K., Tucker, P., Vanhoucke, V., Vasudevan, V., Viegas, F., Vinyals, O., Warden, P., Wattenberg, M., Wicke, M., Yu, Y. & Zheng, X. (eds) (2016). TensorFlow: Large-Scale Machine Learning on Heterogeneous Systems. http://tensorflow.org/ (accessed 30.04.2024).

AI and Natural History (ed) (2023). Shark Tooth Model Dataset. https://universe.roboflow.com/ai-and-natural-history-d1nwu/shark-tooth-model-x8teb (accessed 30.04.2024).23

Kobak D. & Berens P. (2019). The art of using t-SNE for single-cell transcriptomics. *Nature communications*, 10(1): 5416.

Landini W., Altamirano-Sierra A., Collareta A., Di Celma C., Urbina M. & Bianucci G. (2017a). The late Miocene elasmobranch assemblage from Cerro Colorado (Pisco Formation, Peru). *Journal of South American Earth Sciences*, 73: 168-190.

Landini W., Collareta A., Di Celma C., Malinverno E., Urbina M. & Bianucci G. (2019). The early Miocene elasmobranch assemblage from Zamaca (Chilcatay Formation, Peru). *Journal of South American Earth Sciences*, 91: 352-371.

LeCun Y., Bengio Y. & Hinton G. (2015). Deep learning. *Nature*, 521(7553): 436-444.

Linardatos P., Papastefanopoulos V. & Kotsiantis S. (2020). Explainable AI: A review of machine learning interpretability methods. *Entropy*, 23(1): 18. https://doi.org/10.3390/e23010018

Lipovetsky S. & Conklin M. (2001). Analysis of regression in game theory approach. *Applied stochastic models in business and industry*, 17(4): 319-330.

Liu X. & Song H. (2020). Automatic identification of fossils and abiotic grains during carbonate microfacies analysis using Deep Convolutional Neural Networks. *Sedimentary Geology*, 410: 105790.

Liu X., Jiang S., Wu R., Shu W., Hou J., Sun Y., Sun J., Chu D., Wu Y. & Song H. (2023). Automatic taxonomic identification based on the Fossil Image Dataset (> 415,000 images) and deep convolutional neural networks. *Paleobiology*, 49(1): 1-22.

Lundberg S. (ed) (2018). SHAP software. https://shap.readthedocs.io/en/latest/ (accessed 30/04/2024).

Lundberg S.M. & Lee S.I. (2017). A unified approach to interpreting model predictions. *In* Guyon I., Von Luxburg U., Bengio S., Wallach H., Fergus R., Vishwanathan S. & Garnett R.29

FIGURE CAPTIONS

Fig. 1 - Hypothetical phylogenetic tree of the class Chondrichthyes, showing the relationship between the eight shark superorders (names highlighted in bold and black silhouettes). Tree topology after Díaz-Jaimes et al. (2016).

Fig. 2 - Sample images from our composite dataset, depicting one specimen for each studied genus. Note that the teeth are depicted without a scale bar, in either lingual or labial view, and on a black background.

Fig. 3 - Cross-Validation learning curves of SharkNet-X. Train and Validation accuracy and loss are shown for each fold.

Fig. 4 - SharkNet-X best model Train and Validation accuracy (top) and loss (bottom).

Fig. 5 - Confusion Matrix for the test dataset obtained using the best CNN model.

Fig. 6 - t-SNE 2D visualization of the features extracted from the last dense layer of SharkNet-X using the entire dataset.

Fig. 7 - Each row shows a heatmap of SHAP values plotted against the corresponding photographs for sample images of all genera in the shark tooth dataset, with each column corresponding to a specific class that the model predicts. The number of columns is equal to the number of classes because this particular plot illustrates the contribution of each feature to the model's prediction for each class. Therefore, by visualizing SHAP values (represented by blue and red pixels) across different instances and classes, this visualization method supports the understanding of how individual features influence the model's decision-making process and contribute to predictions across different categories.

Fig. 8 – Close-ups of the SHAP heatmaps for 3 sample images in our dataset, showing the original image (left column) besides four relevant heatmaps.

TABLE CAPTIONS

Tab. 1 - Overview of the (sub)datasets of shark tooth images for classification, with indication of the specimen repository for the two private (sub)datasets.

Tab. 2 - Composition of the final, composite dataset of shark tooth images used to train SharkNet-X.

Tab. 3 - SharkNet-X Convolutional Neural Network architecture. For each layer composing the CNN, it shows the type, output shape and number of parameters. At the bottom of the table, a summary of the total number of parameters is provided. 'Trainable' and 'Non-Trainable' refer to the possibility of adjusting the network parameters during the training process.

Tab. 4 – Five-fold Cross-validation performances on Validation and Test set obtained by SharkNet-X.

Tab. 5 - Best model prediction metrics on the Test set obtained by SharkNet-X.

# TABLES

[*Possible positions for the four tables: Table 1 within paragraph 3; Table 2 within paragraph 3; Table 3 within paragraph 4; Table 4 within paragraph 5*]

| Name | Number of images | Reference |
|---|---|---|
| Shark Tooth Model Computer Vision Project | 700 | Shark Tooth Model Dataset - smcm ai class |
| Roboflow-100, shark teeth Computer Vision Project | 280 | shark teeth Dataset - Robotflow 100 |
| AI and Natural History - Shark Tooth Model Computer Vision Project | 115 | Shark Tooth Model Dataset - AI and Natural History |
| MHS Data - Shark Tooth Data Computer Vision Project | 46 | Shark Tooth Data Dataset - MHS Data |
| Megalodon or Not Megalodon Computer Vision Project | 41 | Megalodon or Not Megalodon Dataset - MHS Data |
| Shark sorting Computer Vision Project | 40 | Shark sorting Dataset - Shark sorting |
| GAMPS (Casati-Zanaga collection of Tuscan Pliocene shark teeth) | 361 | Not available - private dataset |
| Departamento de Paleontología de Vertebrados MUSM (collection of Peruvian Miocene shark teeth) | 140 | Not available - private dataset |

**Table 1**

| Genus | Number of images (train/test) |
|---|---|
| *Carcharhinus* | 267/26 |
| *Carcharias* | 233/22 |
| *Carcharocles* | 101/7 |
| *Chlamydoselachus* | 14/3 |
| *Cosmopolitodus* | 72/7 |
| *Galeocerdo* | 128/11 |
| *Hemipristis* | 209/22 |
| *Notorynchus* | 109/11 |
| *Prionace* | 38/5 |
| *Squatina* | 20/4 |
| Total number of images | 1309 (1191/118) |

**Table 2**

| Layer (type) | Output Shape | Number of Parameters |
|---|---|---|
| conv2d (Conv2D) | (None, 222, 222, 32) | 896 |
| max_pooling2d (MaxPooling2D) | (None, 111, 111, 32) | 0 |
| conv2d_1 (Conv2D) | (None, 109, 109, 32) | 9.248 |
| max_pooling2d_1 (MaxPooling2D) | (None, 54, 54, 32) | 0 |
| conv2d_2 (Conv2D) | (None, 52, 52, 64) | 18.496 |
| max_pooling2d_2 (MaxPooling2D) | (None, 26, 26, 64) | 0 |
| flatten (Flatten) | (None, 43264) | 0 |
| dense (Dense) | (None, 128) | 5.537.920 |
| dropout (Dropout) | (None, 128) | 0 |
| dense_1 (Dense) | (None, 10) | 1.290 |
| Total parameters: 5.567.850　Trainable parameters: 5.567.850　Non-Trainable parameters: 0 | | |

**Table 3**

| CV fold | Validation | Test |
|---|---|---|
| 1 | 84.81 | 87.04 |
| 2 | 84.32 | 87.96 |
| 3 | 83.05 | 80.55 |
| 4 | 81.78 | 84.26 |
| 5 | 83.475 | 84.26 |
| **Overall** | **83.5% (std 1%)** | **84.8% (std 2.6%)** |

**Table 4**

| Accuracy | Balanced Accuracy | Precision | Recall | f1 |
|---|---|---|---|---|
| 0.88 | 0.83 | 0.89 | 0.88 | 0.87 |

**Table 5**

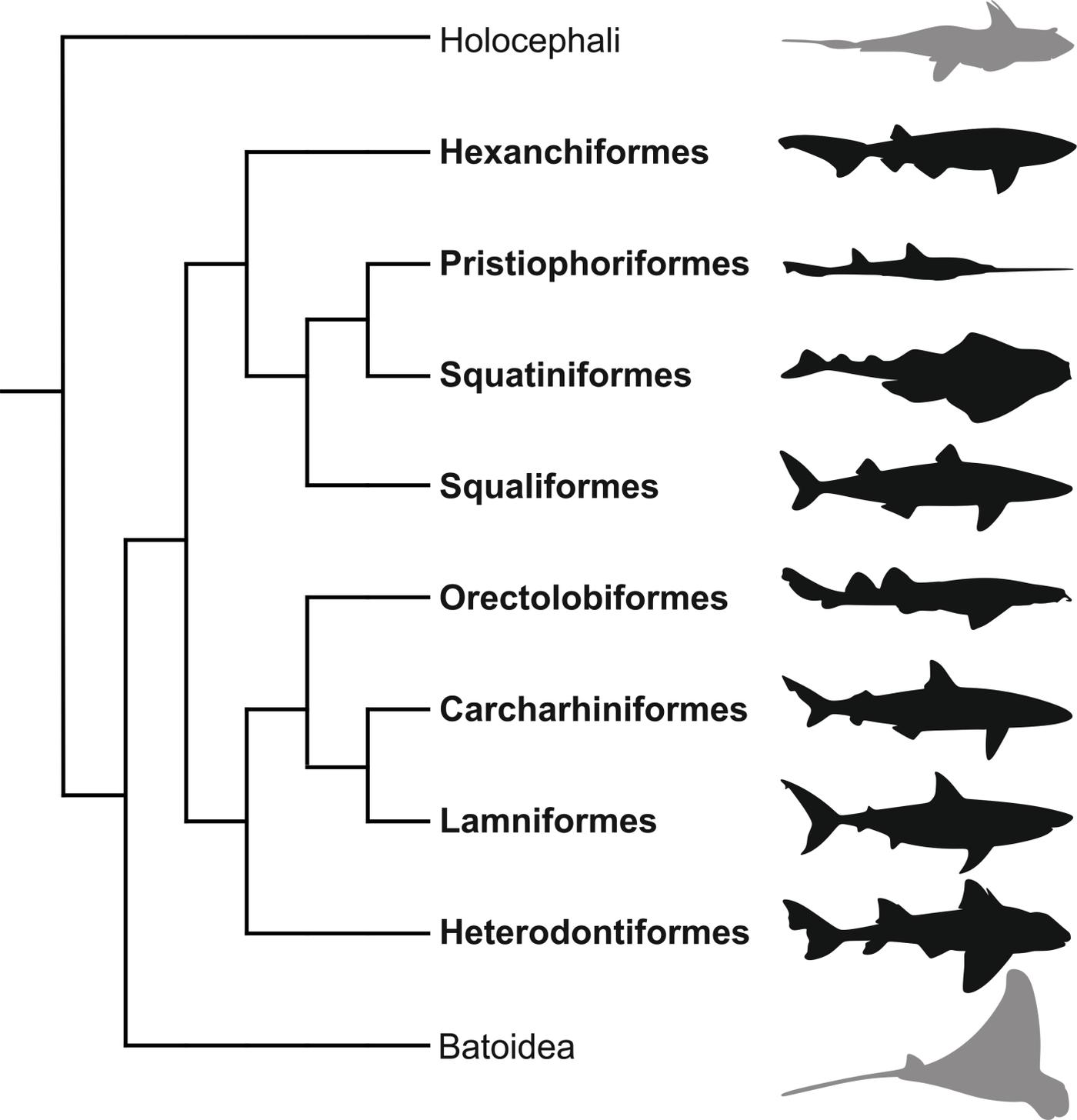

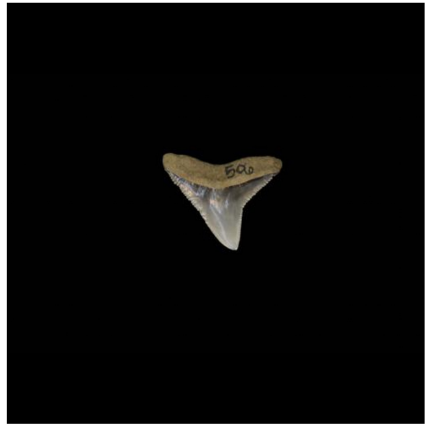 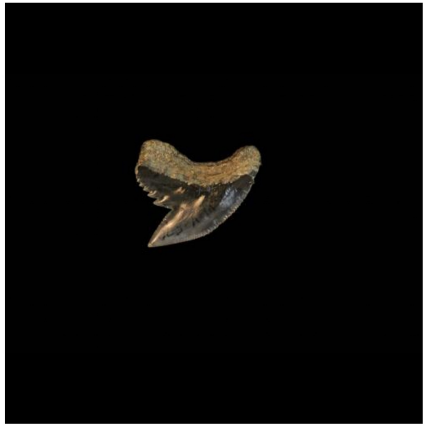 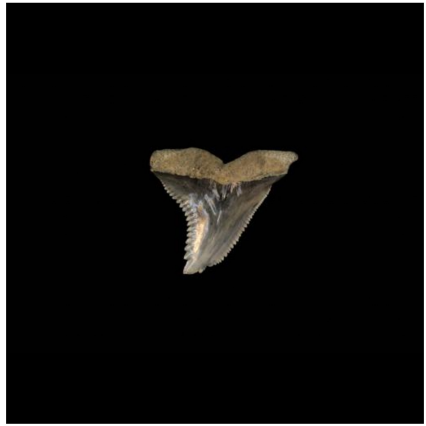 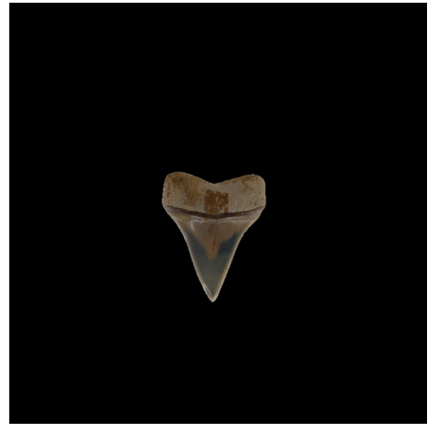 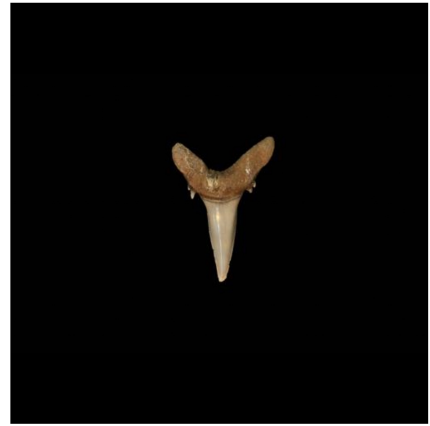 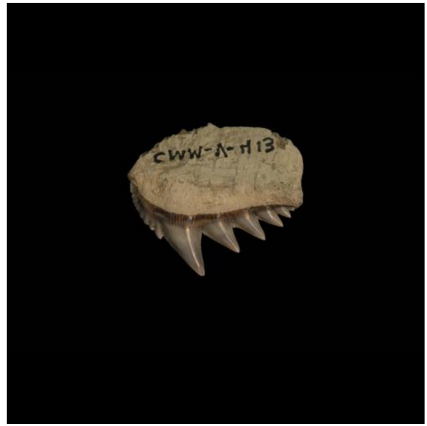 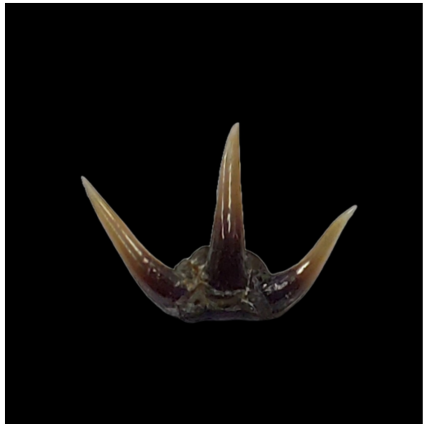 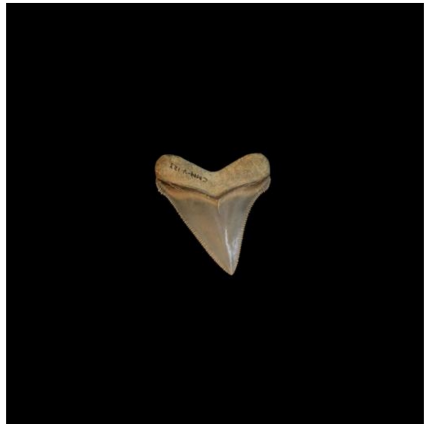 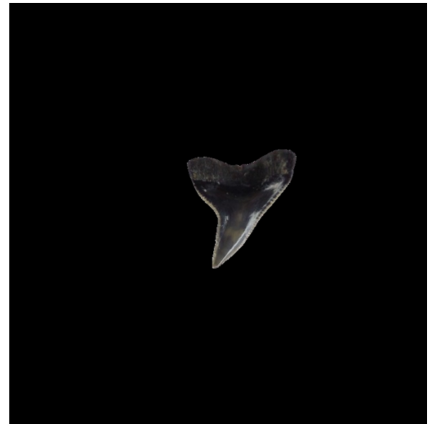 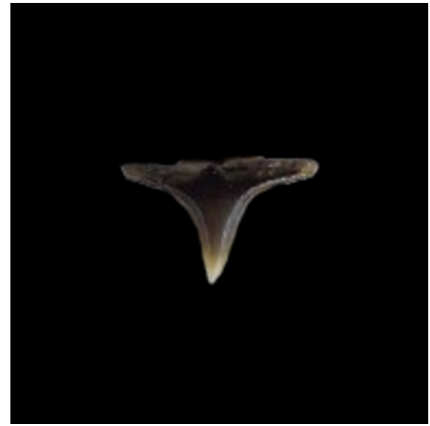

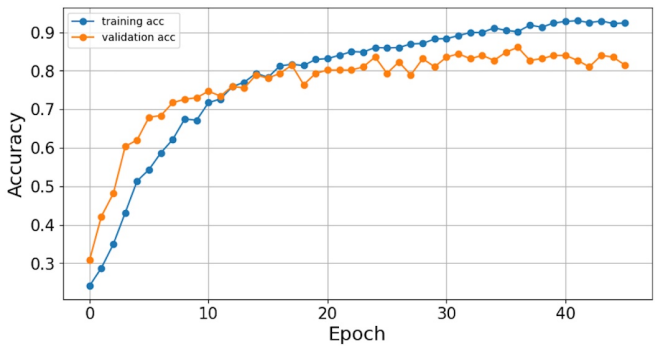
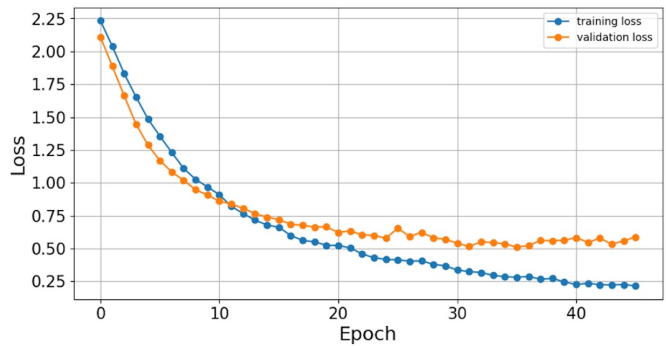
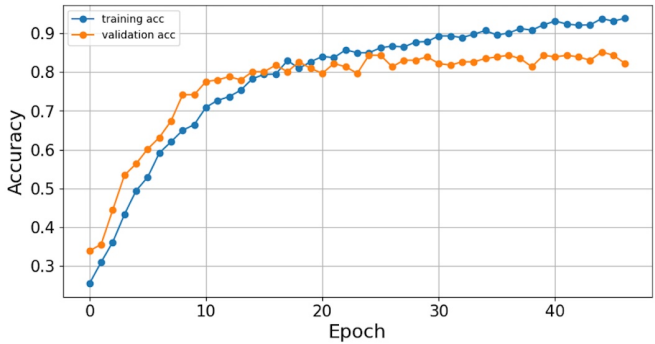
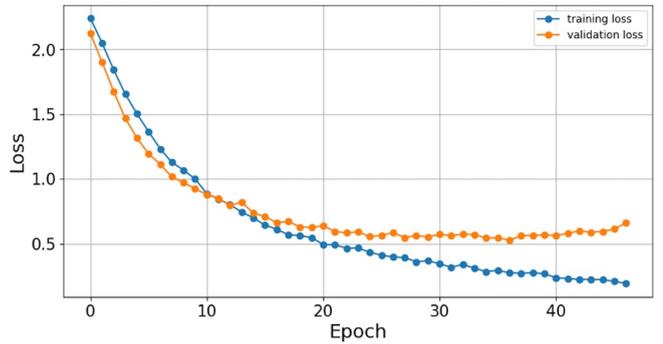
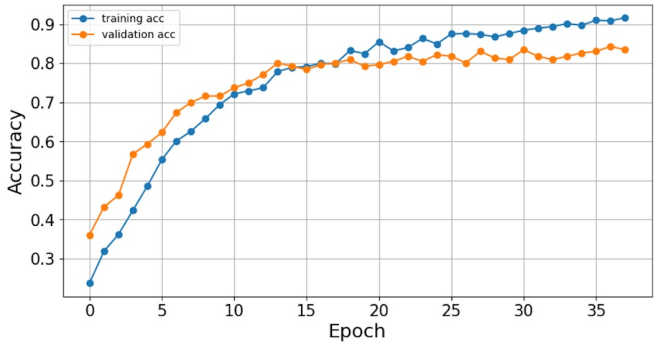
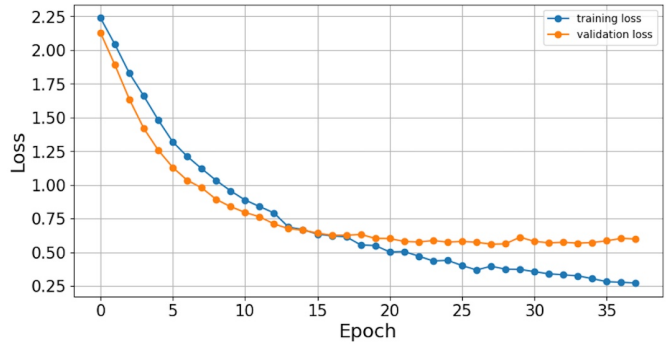
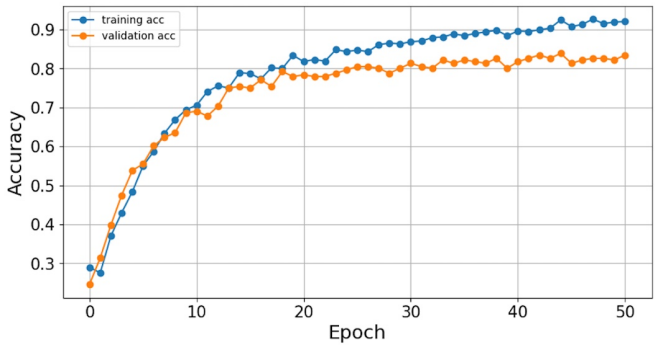
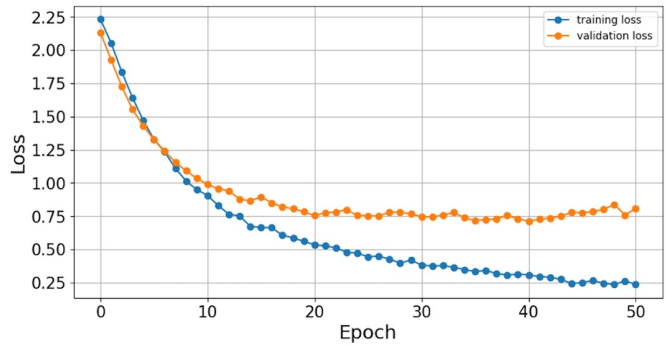
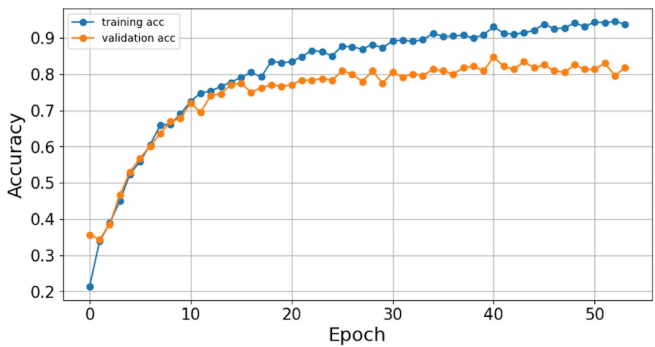
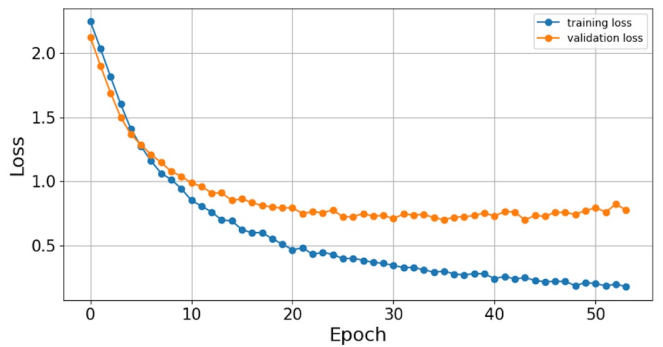

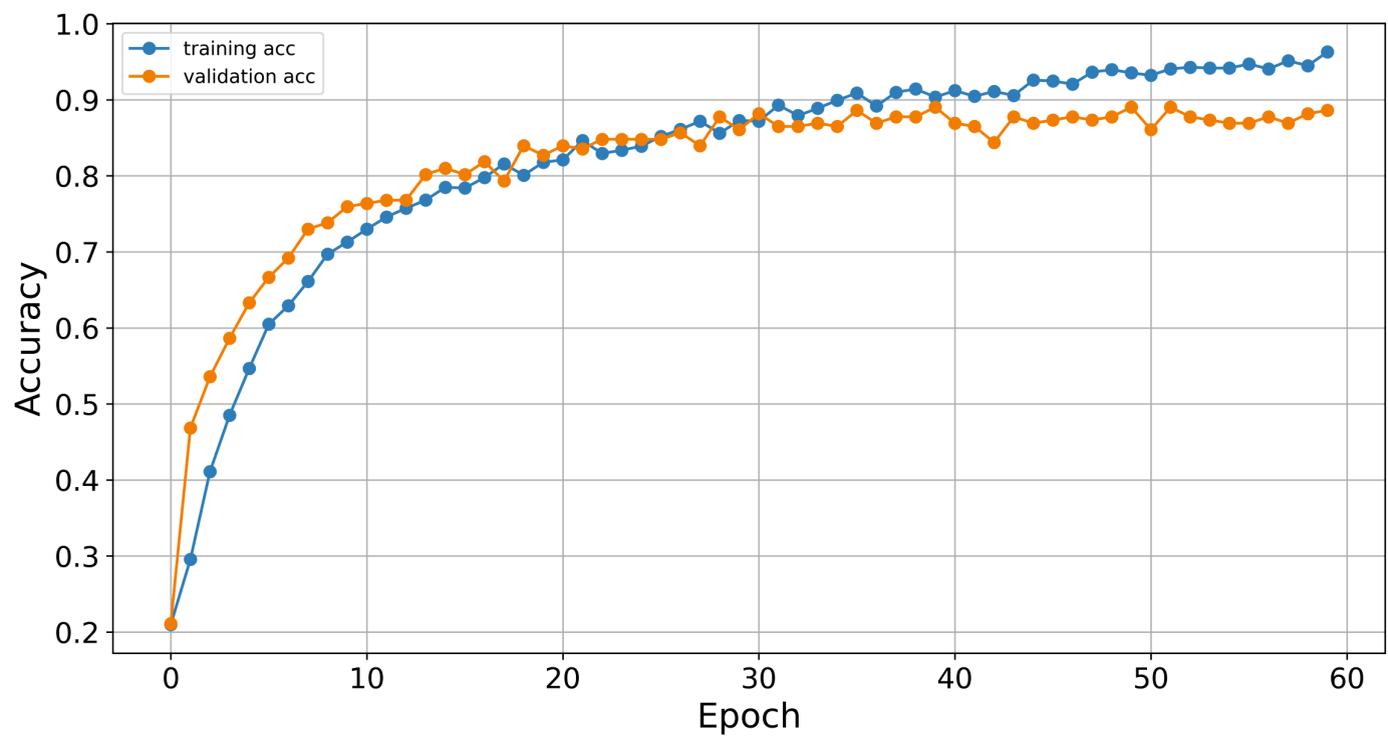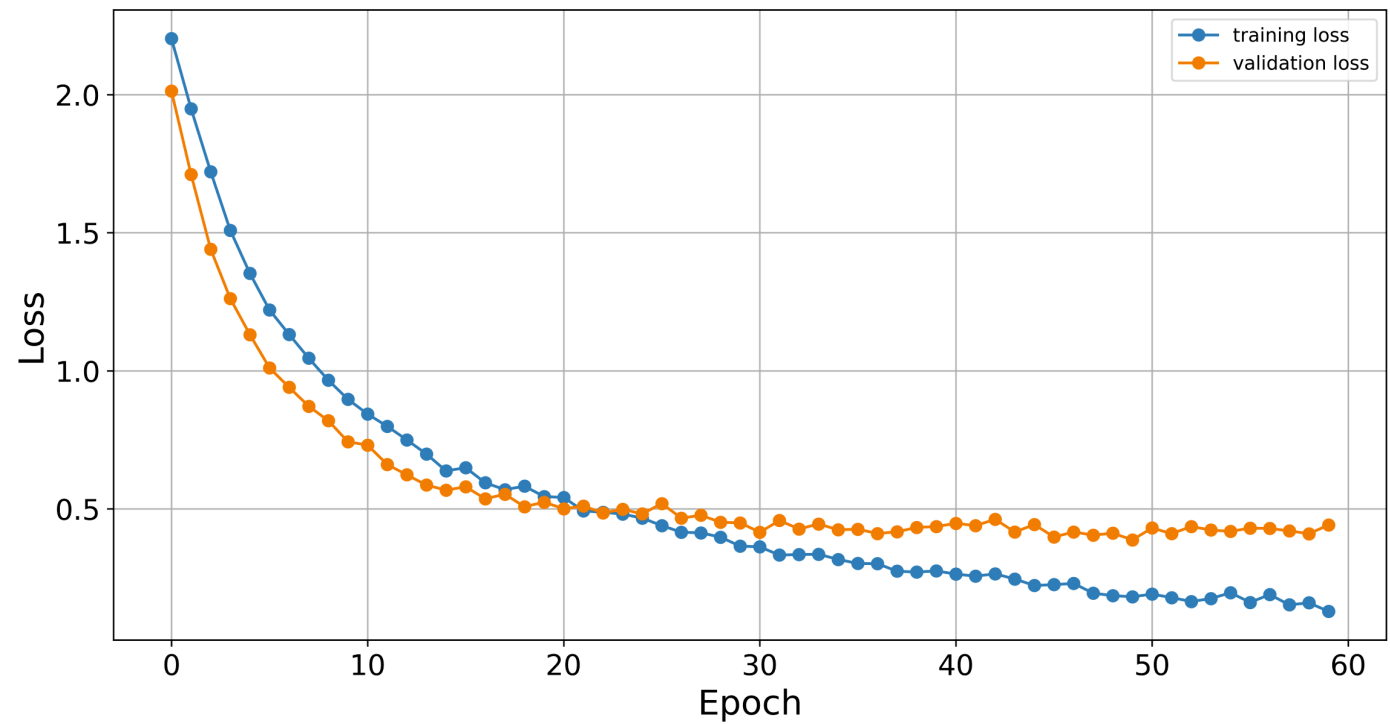

|  | Carcharhinus | Carcharias | Carcharocles | Chlamydoselachus | Cosmopolitodus | Galeocerdo | Hemipristis | Notorynchus | Prionace | Squatina |
|---|---|---|---|---|---|---|---|---|---|---|
| Carcharhinus | 22 | 0 | 0 | 0 | 0 | 1 | 2 | 0 | 0 | 0 |
| Carcharias | 0 | 20 | 0 | 0 | 0 | 0 | 1 | 0 | 0 | 0 |
| Carcharocles | 1 | 0 | 5 | 0 | 0 | 0 | 0 | 0 | 0 | 0 |
| Chlamydoselachus | 0 | 0 | 0 | 2 | 0 | 0 | 0 | 0 | 0 | 0 |
| Cosmopolitodus | 2 | 1 | 0 | 0 | 3 | 0 | 0 | 0 | 0 | 0 |
| Galeocerdo | 0 | 0 | 0 | 0 | 0 | 8 | 1 | 1 | 0 | 0 |
| Hemipristis | 0 | 0 | 0 | 0 | 0 | 0 | 21 | 0 | 0 | 0 |
| Notorynchus | 0 | 0 | 0 | 0 | 0 | 1 | 0 | 9 | 0 | 0 |
| Prionace | 1 | 0 | 0 | 0 | 0 | 0 | 0 | 0 | 3 | 0 |
| Squatina | 0 | 1 | 0 | 0 | 0 | 0 | 0 | 0 | 0 | 2 |

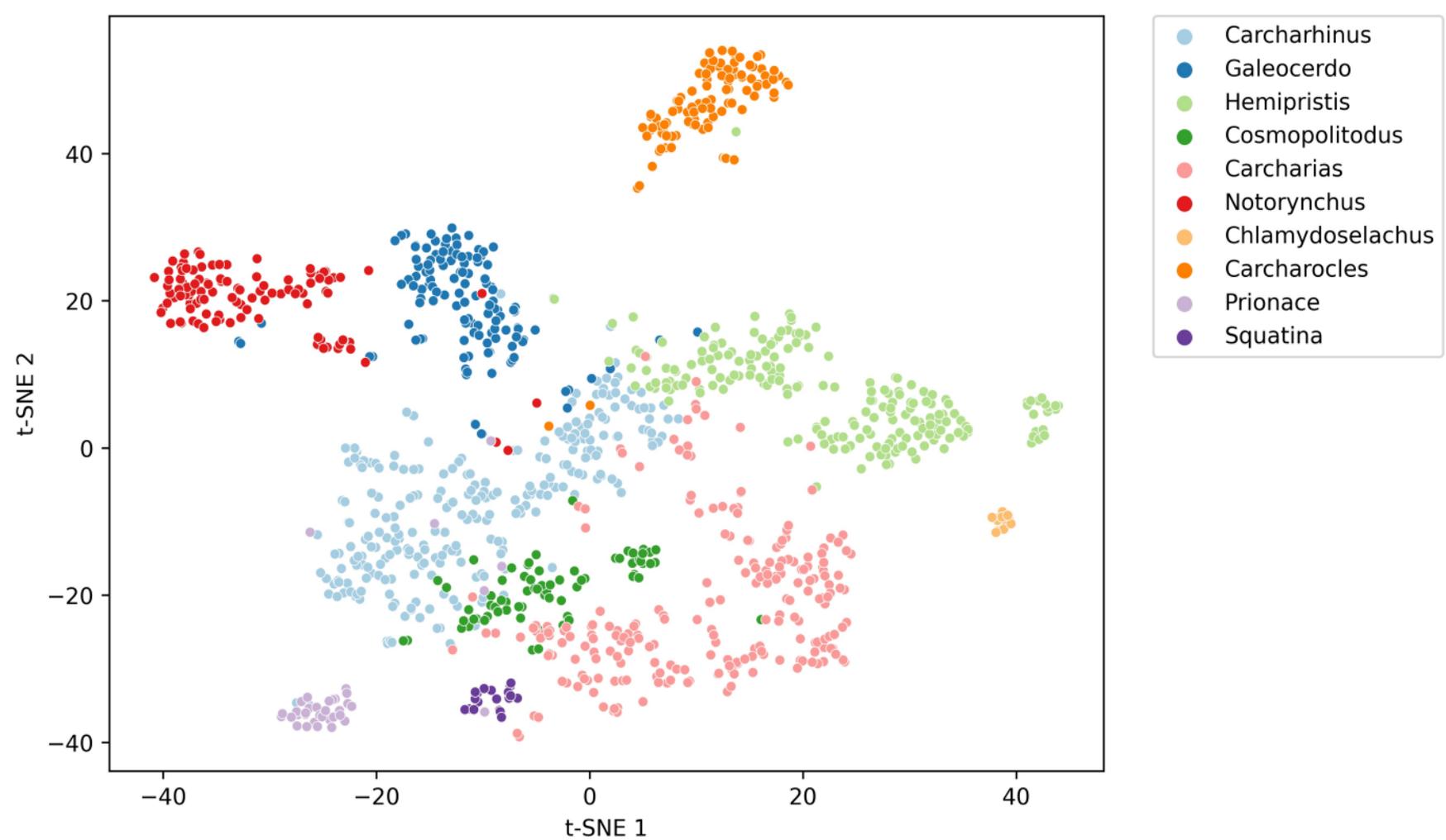

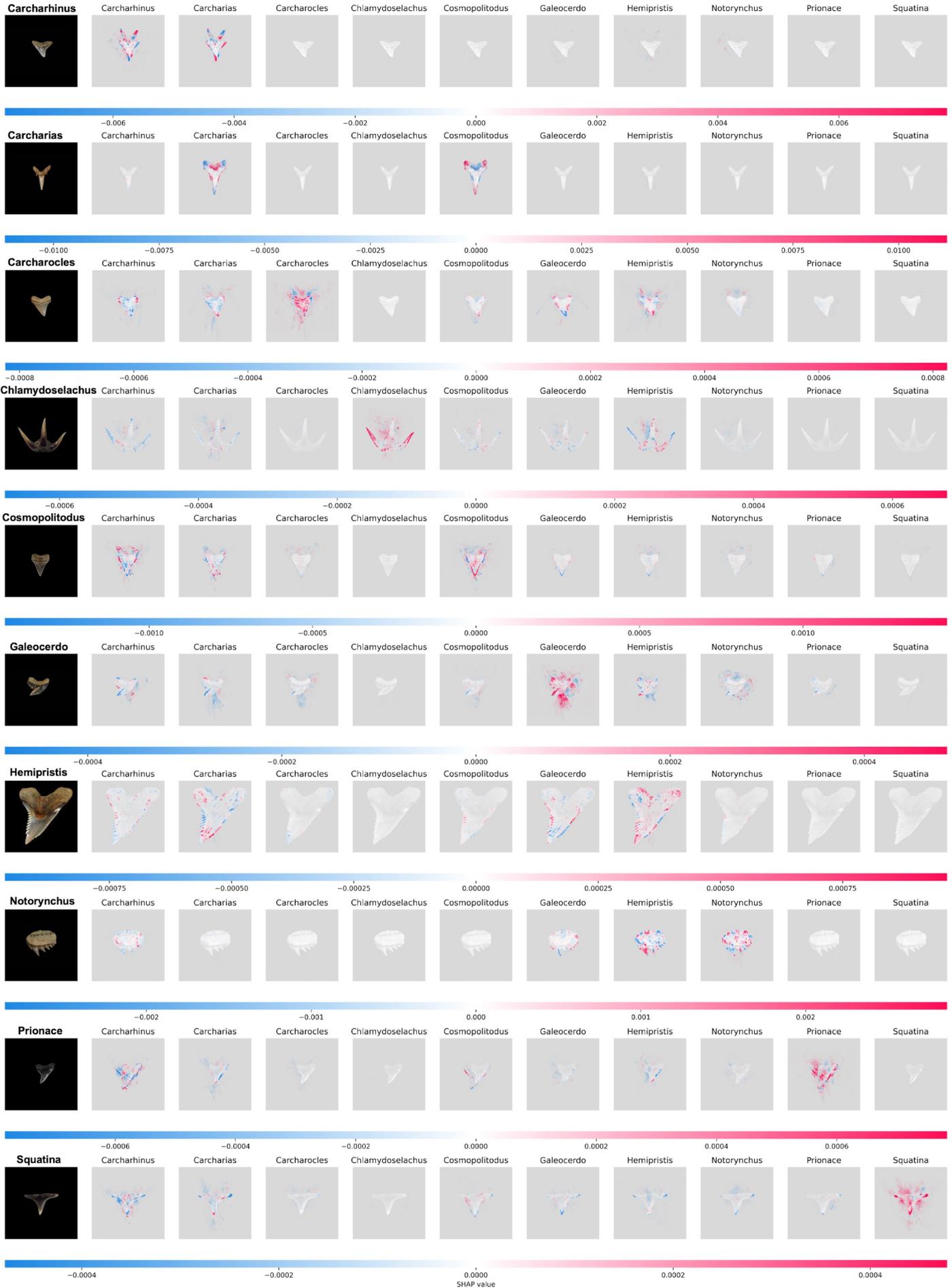

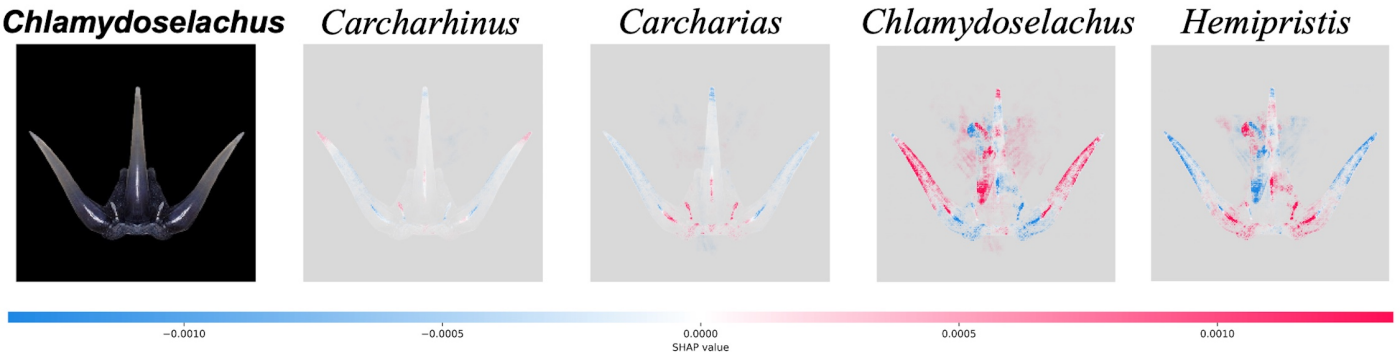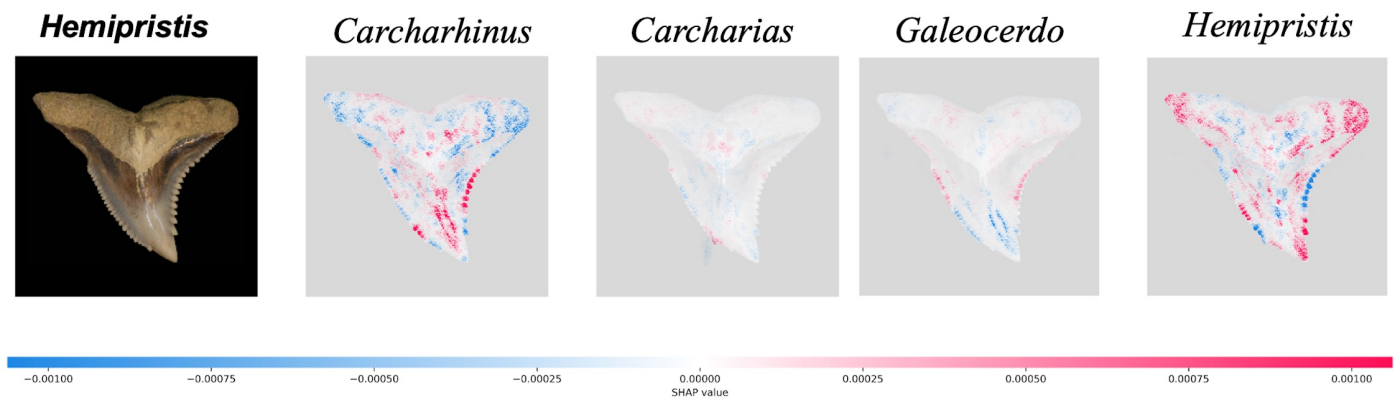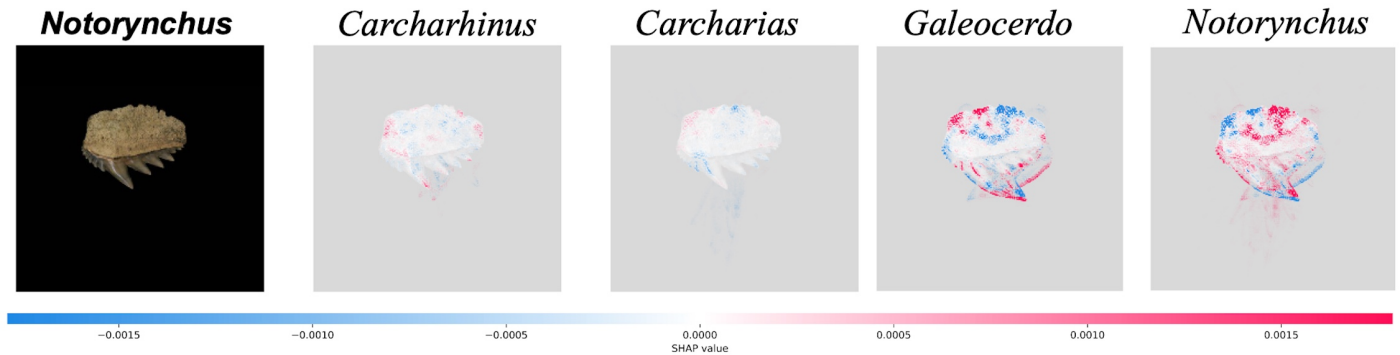